\documentclass[english]{article}
\usepackage[utf8]{inputenc}
\usepackage[T1]{fontenc}
\usepackage{babel}
\usepackage{amsmath}
\usepackage{graphicx}
\usepackage{subcaption}
\usepackage{fancyhdr}
\usepackage{csquotes}
\usepackage{amstext}
\usepackage{graphicx}
\usepackage[natbib=true,style=numeric,sorting=none]{biblatex}

\begin{document}

\title{A Novel Sentiment Analysis Engine for Preliminary Depression Status Estimation on Social Media}

\author{Sudhir Kumar Suman\textsuperscript{2}, Hrithwik Shalu\textsuperscript{3}, Lakshya A Agrawal\textsuperscript{4},\\ Archit Agrawal\textsuperscript{4}, Juned Kadiwala\textsuperscript{1{*}}}

\maketitle
\thispagestyle{fancy}
\noindent
1. University of Cambridge     
\\
2. Indian Institute of Technology Bombay
\\
3. Indian Institute of Technology Madras
\\
4. Indraprastha Institute of Information Technology, Delhi
\\
{*}corresponding author

\begin{abstract}
Text sentiment analysis for preliminary depression status estimation of users on social media is a widely exercised and feasible method, However, the immense variety of users accessing the social media websites and their ample mix of vocabularies makes it difficult for commonly applied deep learning-based classifiers to perform. To add to the situation, the lack of adaptability of traditional supervised machine learning could hurt at many levels. We propose a cloud-based smartphone application, with a deep learning-based backend to primarily perform depression detection on Twitter social media. The backend model consists of a RoBERTa based siamese sentence classifier that compares a given tweet (Query) with a labeled set of tweets with known sentiment ( Standard Corpus ). The standard corpus is varied over time with expert opinion so as to improve the model's reliability. A psychologist ( with the patient's permission ) could leverage the application to assess the patient's depression status prior to counseling, which provides better insight into the mental health status of a patient. In addition, to the same, the psychologist could be referred to cases of similar characteristics, which could in turn help in more effective treatment. We evaluate our backend model after fine-tuning it on a publicly available dataset. The find tuned model is made to predict depression on a large set of tweet samples with random noise factors. The model achieved pinnacle results, with a testing accuracy of 87.23\% and an AUC of 0.8621. We believe that the problem of lack of adaptability of a natural language model to estimate mental health status in a large and vibrant social media community such as Twitter could be effectively solved by introducing the proposed application into the society.
\end{abstract}

\section*{Introduction}
Depression is one of the most common mental health disorders that today's society suffers from, with more than 264 million affected individuals\cite{who_depression}. Out of the tens of millions of people who suffer from depression every year, only a small percentage receive adequate medical treatment\cite{inproceedings}. It has a profound effect on every aspect of an individual's life. Depression leads to reduced interest and productivity in a person's day to day activities. Due to the same, it has become the leading cause of disability around the globe. Depression severely impacts individuals' ability to cope with daily life situations and therefore differs drastically from normal mood variations experienced by many. It can even lead to suicide. As per WHO estimates \cite{who_depression}, in the year 2015, 788,000 people have died by suicide and that it was the second largest  cause ( common ) of death for people aged between 15 and 29 years old worldwide. In addition to this, depression has been a core cause of several physical conditions, such as tuberculosis or cardiovascular diseases\cite{Whooley2008}. Having analyzed the above facts, it could be confirmed that depression status estimation is of significant importance in ensuring mental-well being in today's society.

Proper treatment administration to people suffering from depression is often missed due to the limited availability of mental health care, especially in conflict regions, apart from personal reasons\cite{paper2}. Effective preliminary diagnosis is one methodology that could provide optimized allocation of mental health services. About 75\% of the college students that need such mental health care do not seek the service\cite{HUNT20103}. Depression and other mental illnesses have been a core reason for social withdrawal and isolation. Modern-day society is at large influenced by the presence of social media networks. Today social media websites are the largest source of information regarding anything and anyone in the whole world. It was found that social media platforms are indeed increasingly used by affected individuals to connect with others, share experiences, and support each other. Overcoming stigma problems has been traditionally considered a significant benefit of online platforms and the recently introduced mobile mental health interventions. A localized study \cite{berger2005internet} ultimately provided concrete proof of the fact that users in social-media with stigmatized illnesses like depression are more inclined to make use of online resources for health-related information and communication about their condition. Efficient ordering of patients in queue is a deciding factor in increasing healthcare efficiency, to which preliminary analysis methods play a key role. Preliminary analysis of depression through user data extraction from social media is a proven and effective methodology. With the above facts one could conclude that an effective means to provide effective treatment for depression in our large and vibrant community is by analyzing social media conversations performed by individuals. Because of the promise and feasibility of the same, it has become an active area of research in psychology and computation.

Methods from the regime of Natural language processing (NLP) are at large deployed to help with the cause in the modern day world. Recent advancements in the field ( such as \cite{vaswani2017attention} ) has proven effectiveness in tackling many complex natural language analysis tasks such as machine translation, sentiment analysis etc. But the tremendous variation of vocabulary amongst social media users poses a huge challenge for language models employed for specific tasks to perform effectively. We devised a smartphone application based classification methodology which utilizes siamese Sentence-BERT \cite{reimers2019sentencebert} networks for comparative analysis of social media conversations, rather than unique feature recognition based classification. With the proposed methodology we hope to achieve state of the art results in the task of depression status prediction and at the same time ensure optimal patient ordering and management, so as to increase the efficiency of the mental health care services offered in today's modern day world.

\section{Related Work}
Social network analysis for depression status estimation has been a widely investigated topic in the modern day world, Some of the early approaches for analysing depression were questionnaire based rating. Several measures such as  CES-D \cite{doi:10.1177/014662167700100306}, BDI \cite{Beck1961AnIF}, and SDS \cite{10.1001/archpsyc.1965.01730060026004}  were proposed and were proven to be effective. A work \cite{dechoudhury2013predicting} created much awareness on why social media acts as a great platform for depression status estimation. In a work \cite{10.1145/2702123.2702280} the effectiveness of a merged approach with social media activity along with questionnaire response was investigated, they concluded that large observation periods ( > 2 months ) in user activity for current depression estimation lead to worsening the model's accuracy. \cite{nadeem2016identifying} formulated the task of Major Depressive Disorder (MDD) analysis as a text classification problem solved using statistical methods and were able to achieve a peak accuracy of 81\% on their data. Neural network based classifiers were utilized for early stage depression detection in \cite{8580405}.  Having inspired by the same many further studies had formulated the problem in a similar manner using state of the art natural language models (NLP). Architectures such as Bidirectional LSTM's were employed in many studies, in \cite{Chen_2018} a unified approach with varied embedding representation and BiLSTM's with attention mechanism were used to analyse tweet sentiment. \cite{reddy2019twitter} carried out sentiment analysis through a novel distributed preprocessing method. Advancements in NLP such as the introduction of attention based architectures such as the Transformers \cite{vaswani2017attention} caused a surge in related studies. BERT \cite{devlin2019bert} based pretraining showed improved performance in many natural language inference tasks. RoBERTa model \cite{liu2019roberta} further optimized on BERT based pretraining to achieve state of the art results in many NLP tasks. In a recent study \cite{murarka2020detection} a RoBERTa based classifier was employed for the task of twitter sentiment analysis and achieved pinnacle results. Having inspired from the same and with the understanding of the limitations of the literature, our backend model is based on the idea of generating meaningful sentence representations using the Sentence-BERT model \cite{reimers2019sentencebert} for textual comparison.

\section{Methods}
The proposed methodology for depression detection from social media primarily consists of the front-end smartphone application which a user ( being a psychologist/psychiatrist ) could use to gain insight on a patient prior to detailed diagnosis. The application built to conduct the study was build using default UI/UX material design from android studio. The backend model was deployed in google cloud platform (GCP). An in house API was build to communicate between the social media and the android client, once authentication approval requests are obtained. The pre-build twitter API is used to extract relevant public information from the user's account to send a query to the cloud backend. For local app information (authentication) storage Firebase was utilized.

The backend architecture consisting of Siamese Sentence-BERT was shortly trained on a publicly available twitter sentiment dataset prior to deployment in order to tune the sentence representations. A human allied methodology improving mechanism is build into the app so that the prediction of the backend model could imporve and adapt with time, through the effective collaboration of expert clinicians from around the globe. 

We believe that through the introduced live methodology improvement and flexibility of our backend model, the prominent problems faced by commonly deployed models such as, the lack of adaptability to the enormous social media culture and inefficient re-training could be effectively tackled. Through an effective and easy preliminary patient status estimate a clinician could effectively order the patients for optimized treatment.

\subsection{Application Usage Overview}
\begin{figure}[ht]
\includegraphics[width=12cm, height=10cm]{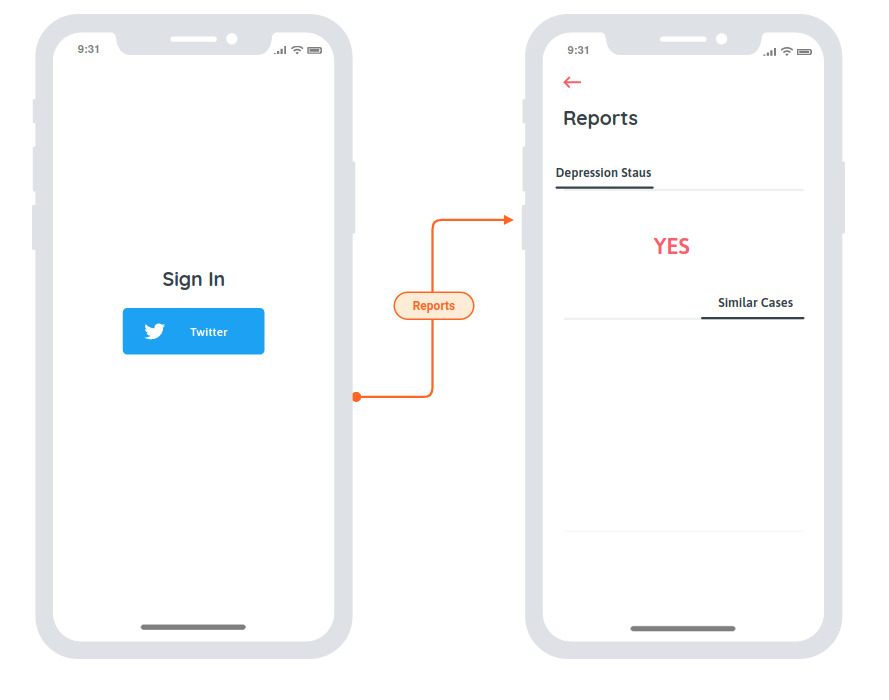}
\caption{Simplicity of the proposed methodology during inference process}
\label{c1}
\end{figure}

The application acts as a pathway between the backend model and the medical practitioner. A clinician with the patient's permission could utilize our methodology by enabling them to provide public profile access by logging in with their social media (such as twitter) credentials. The simplistic overview of the app usage in this manner is conveyed in (Figure \ref{c1}). The application after being provided access to the public profile could collect relevant openly available information (such as public tweets) and send an anonymous query to the cloud backend, The cloud backend model compares the provided query with known cases ( Provided by expert clinicians ) to diagnose the patients depression status ( Update of the known cases for robust diagnosis is key in our methodology and is discussed in a coming section ). Once the inference process is complete the backend reports back a bunch of useful inferences such as current depression status, a few similar cases ( anonymous ) with relevant information ( including treatment history ) providing immense amount of insight to the user (clinician). 

\pagebreak

\subsection{Backend Model}
The key function of our backend model is to form meaningful sentence representations and effectively compare them.
Siamese Networks are widely deployed in representation comparison tasks, the commonly applied loss is the contrastive energy function. For our purposes we chose the loss function for any pair of data points ($x_1$,  $x_2$) as

\begin{equation}
\mathcal{L}(x_1, x_2) = \frac{1}{2}( 1 - y )D^2 + \frac{1}{2}max\{0 , m - D\}^2
\label{e1}
\end{equation}

\noindent
Where, margin parameter ($m$) : desired $D$($x_1$,  $x_2$) for unlike pairs

\begin{equation}
D(x_1, x_2) = \left\|W(x_1) - W(x_2) \right\|_2
\end{equation}

\begin{equation}
    y = \begin{cases}
        0 \text{, if ($x_1$,  $x_2$) are of same class} \\
        1 \text{, if ($x_1$,  $x_2$) are of different class} \text{.}
    \end{cases}
\label{e3}
\end{equation}

\pagebreak

\begin{figure}[h!]
\includegraphics[width=12cm, height=15cm]{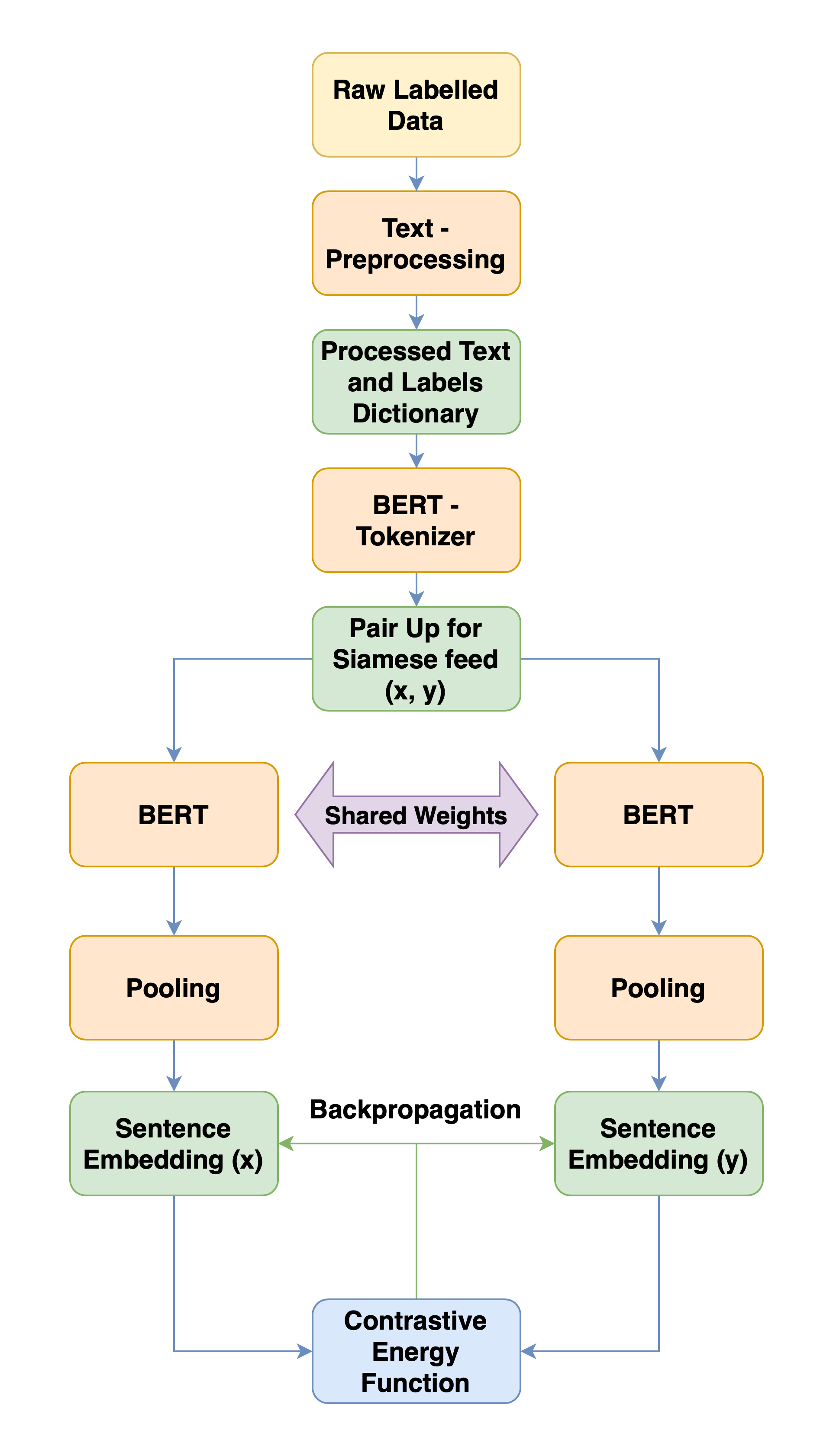}
\caption{Training process of the Siamese Sentence-BERT network visualized}
\label{c2}
\end{figure}

\begin{figure}[h!]
\includegraphics[width=12cm, height=13cm]{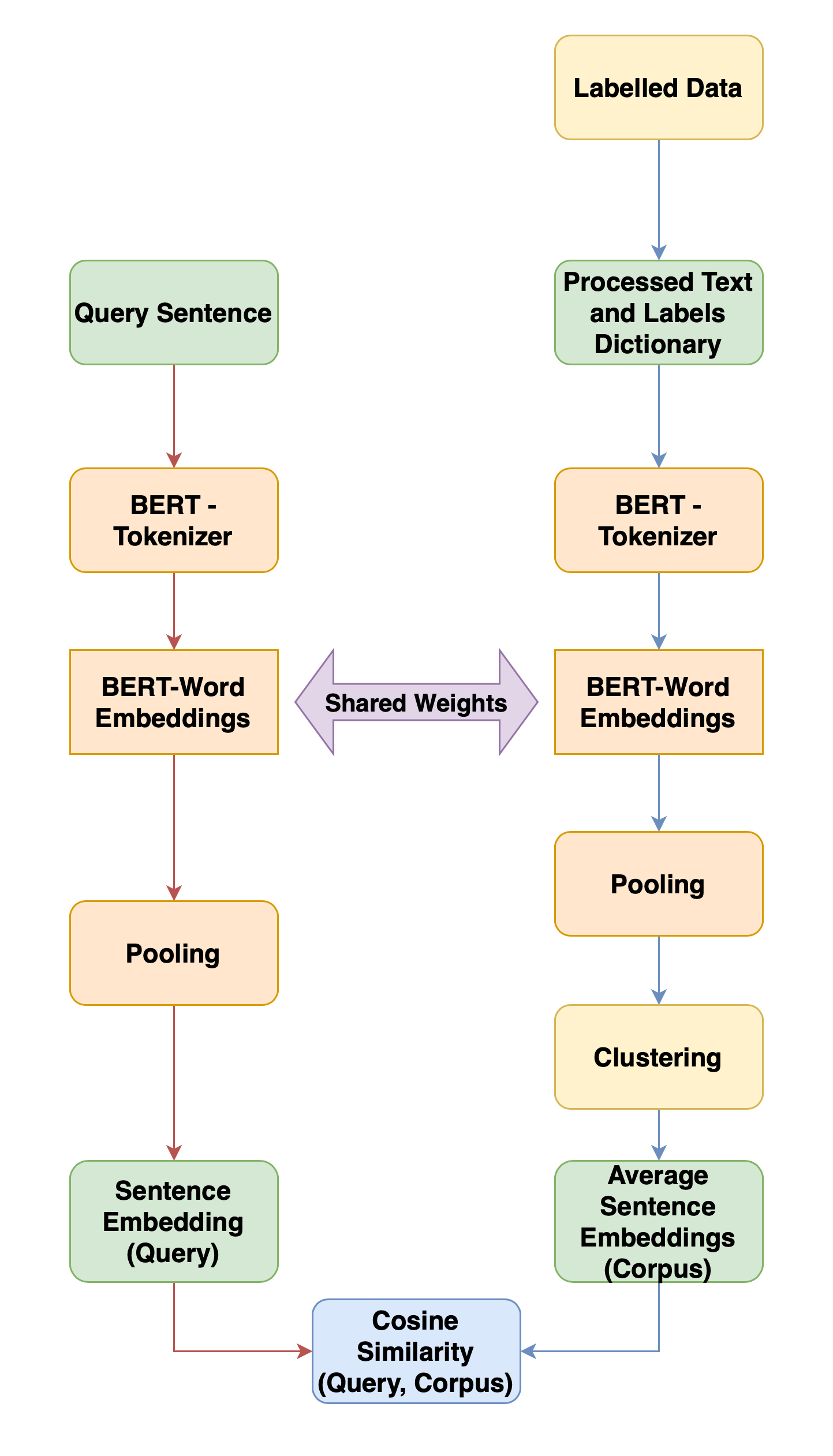}
\caption{Inference process for a processed general query sentence}
\label{c3}
\end{figure}

To tune the pre-trained RoBERTa network, we first process the text from the public dataset to remove unwanted artifacts ( such as URLs ), to avoid unforeseen errors in prediction. Once pre-processed the clean text is tokenized and paired up, with binary labels as per equation \ref{e3}. The formed pairs of data are used to tune the Sentence-BERT model with the loss estimate is calculated as per equation \ref{e1}. Once tuned the same network is used to generate corresponding sentence embedding for a standard set of tweets. The vector set of embeddings so formed are clustered using K-Means clustering mechanism. The individual cluster averages are used as an effective and fast inference standard. The inference process is highlighted in (Figure \ref{c3}). The similarity between the query ( $q$ ) and a class average embedding ( $A_i$ ) is calculated as 

\begin{equation}
    cos(\pmb q, \pmb A_i) = \frac {\pmb q \cdot \pmb A_i}{||\pmb q|| \cdot ||\pmb A_i||}
\label{e4}
\end{equation}

The appropriate class is selected as per maximum similarity criteria.\\ ( Note : the standard set of tweets are initially selected with expert advise )

\subsection{Human Allied Improvement}
With more than 3.81 billion users worldwide, the social media is truly an ocean of vibrant cultures. The extensive variation in user language vocabulary due to use of localised language slang, social media language alteration with time etc makes it impossible for a deployed model to maintain effective and continued performance. To tackle the same we use a human allied methodology improvement mechanism that is made possible by the vibrant worldwide user ( clinician ) community. Experts from around the globe could contribute in providing relevant data ( such as tweets or public comments ) from locally diagnosed cases. The backend model could in turn use this information for re-estimating the average sentence embedding of each available class, thus enforcing the effective comparisons to imporve. Re-clustering occurs when sufficient data limit is ac hived and extremely similar cases are disposed off through a semantic search ( terminates once threshold similarity is achieved).

\begin{figure}[h!]
\includegraphics[width=12cm, height=11cm]{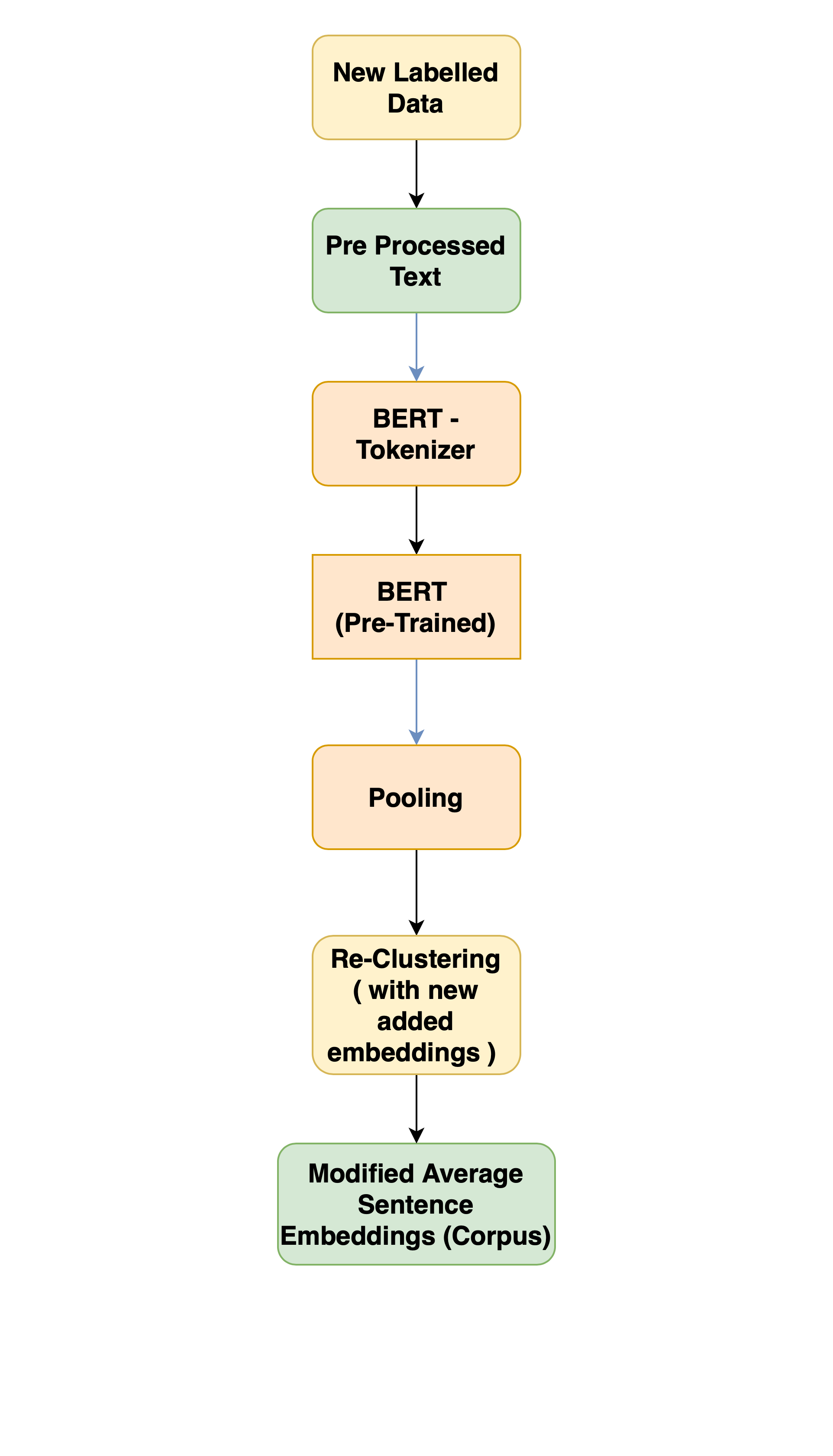}
\caption{The backend part of the human allied improvement visualized}
\label{c4}
\end{figure}

\section{Experiments and Results}

For our purposes we tune our backend model initially by training the same on the publicly available twitter sentiment dataset \cite{Twitter_data} by the process illustrated in Figure \ref{c2}, The model is then rigorously tested on the testing set to evaluate it's performance. Dataset description and obtained results are as detailed below.

\subsection{Dataset Used}

\begin{figure}[h!]
\includegraphics[width=12cm, height=6cm]{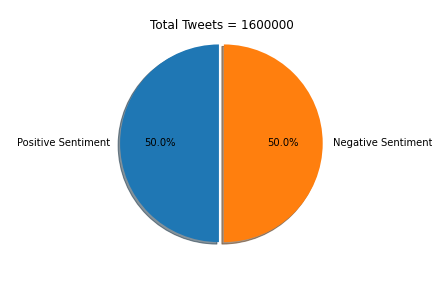}
\caption{Initial class distribution of dataset used \cite{Twitter_data} (Graphical representation)}
\label{d1}
\end{figure}

The dataset used in this study is made publicly available. Rigorous text cleansing methods were applied prior to training and inference in order to ensure maximum compatibility with pretrained BERT models used. For our initial model testing purposes we limit our analysis to binary labels i.e Positive, Negative Sentiments. Out of the available data 10\% tweets from each class was kept apart as testing set, the spilt selection was done randomly. Further information about the initial dataset is as provided in Figure 6.

\subsection{Model benchmarks}

\begin{equation}
    TPR = \frac{TP}{TP + FN}
\end{equation}

\begin{equation}
    FPR = \frac{FP}{FP + TN}
\end{equation}

The dataset used in our study could be the prime focus in many effective research studies yet to come. Our model achieved an accuracy of 87\% with a confidence interval of +/- 0.1855, As an effective benchmark of our model we provide the ROC curve for depression prediction and corresponding AUC. 

\begin{figure}[h!]
\includegraphics[width=12cm, height=8cm]{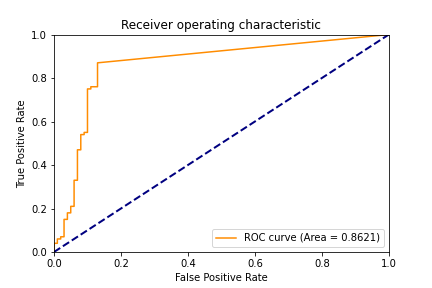}
\caption{ROC Curve for Depression Status Prediction}
\label{mb1}
\end{figure}

\subsection{Comparative Analysis}

The key aspect of our model is it's ability to create meaningful sentence representations for depression sentiment comparisons. To prove the effectiveness of our model we query unknown tweets ( not seen by the model ). The inference process is as conveyed in (Figure \ref{c3}). Once the class of the tweet is selected we run a semantic search algorithm on the standard corpus to determine the most matching tweet to the query. In order to further conclude the robustness of our model we estimate the average similarity value from the negative class.  The similarity index used in our experiment is the standard cosine similarity as per equation \ref{e4}. A few of the results obtained while comparing depressive tweets with the standard set as per our analysis criteria are listed in Table \ref{tab:table1}. With the illustrated results we see that the gap between the top-match depressive tweet and the average similarity of positive sentiment corpus is large.

Note : no spelling correction was done after post processing of the text. The table illustration has anonymous tweets in their initial unprocessed form ( any URL or User information present were removed ) 

\pagebreak

\begin{table}[h]
\centering
\advance\leftskip-3.3cm
\begin{tabular}{llll}
\hline
\textbf{\begin{tabular}[c]{@{}l@{}}Query Tweet \\           ( Negative )\end{tabular}}                                                                                                 & \textbf{\begin{tabular}[c]{@{}l@{}}Top - Match\\            ( Known Sentiment )\end{tabular}}                                                                                            & \textbf{\begin{tabular}[c]{@{}l@{}}Similarity Index\\  ( Top - Match )\end{tabular}} & \textbf{\begin{tabular}[c]{@{}l@{}}Similarity Index Avg.\\  ( Positive Corpus )\end{tabular}} \\ \hline
I wanna go home!                                                                                                                                                                       & I'm so tired of work...I need a life....                                                                                                                                                 & 0.7213                                                                               & 0.1667                                                                                        \\ \hline
\begin{tabular}[c]{@{}l@{}}I'm doing my homework. \\ It's gosh darn hard!!\end{tabular}                                                                                                & \begin{tabular}[c]{@{}l@{}}Why does school take over \\ your life so much you don't get \\ sleep anymore  .. I am still doing \\ school work and have more to do\\  as well\end{tabular} & 0.7189                                                                               & 0.1431                                                                                        \\ \hline
\begin{tabular}[c]{@{}l@{}}It's depressing to start your \\ day knowing you're not coming \\ home until tomorrow night.\end{tabular}                                                   & \begin{tabular}[c]{@{}l@{}}yup night workouts r the worst \\ but unfortunetly my work \\ schedule only allows me to go \\ at night  its tough!\end{tabular}                              & 0.7431                                                                               & 0.1138                                                                                        \\ \hline
\begin{tabular}[c]{@{}l@{}}Feel like I'm stuck in a rut. \\ Waiting to hear back from \\ schools is killing me.  \\ Also decided to take a break \\ from \#wow for awhile\end{tabular} & \begin{tabular}[c]{@{}l@{}}yea it is so quiet around here cuz \\ everyone has to work im bored to \\ death with nobody to talk to\end{tabular}                                           & 0.6495                                                                               & 0.1340                                                                                        \\ \hline
\end{tabular}
\caption{Comparison Chart of Depressive Tweets }
\label{tab:table1}
\end{table}

\section*{Conclusion}
With the obtained results we could ultimately conclude that the methodology used in this study was robust and efficient in the task of classification through latent space comparison. The low confidence interval value points to the precise nature of the obtained testing set accuracy. In today's society social media has become the largest public information source about any individual, which in-turn could be called as the most effective means to assess the mental well-being of a person. But the vast and vibrant culture of the ocean of users in social media makes it a challenge for effective comparison. The proposed methodology to a great extent tackles the above challenge effectively an efficiently. We believe that through the introduction such a methodology globally, the mental-health issues faced by individuals living in today's society could be diagnosed and tackled.

\section*{Code Availability}
The custom Python code and android app used in this study are available from the corresponding author upon reasonable request and is to be used only for educational and research purposes.

\printbibliography

\end{document}